\pdfoutput=1
\documentclass[11pt]{article}
\usepackage{naacl2021}
\usepackage{times}
\usepackage{latexsym}
\usepackage[T1]{fontenc}
\usepackage[utf8]{inputenc}
\usepackage{microtype}
\usepackage{graphicx}
\usepackage{amsmath}
\usepackage{makecell}
\usepackage{amssymb}
\usepackage{ragged2e}

\title{Domain State Tracking for a Simplified Dialogue System}

\author{
    Hyunmin Jeon \\
    Computer Science and Engineering \\
    POSTECH, Pohang, South Korea \\
    \texttt{jhm9507@postech.ac.kr} \And 
    \stepcounter{footnote}Gary Geunbae Lee \\
    Computer Science and Engineering \\
    Graduate School of Artificial Intelligence \\
    POSTECH, Pohang, South Korea \\
    \texttt{gblee@postech.ac.kr} \\
    }

\begin{document}
\maketitle
\begin{abstract}
Task-oriented dialogue systems aim to help users achieve their goals in specific domains. Recent neural dialogue systems use the entire dialogue history for abundant contextual information accumulated over multiple conversational turns. However, the dialogue history becomes increasingly longer as the number of turns increases, thereby increasing memory usage and computational costs. In this paper, we present DoTS (Domain State Tracking for a Simplified Dialogue System), a task-oriented dialogue system that uses a simplified input context instead of the entire dialogue history. However, neglecting the dialogue history can result in a loss of contextual information from previous conversational turns. To address this issue, DoTS tracks the domain state in addition to the belief state and uses it for the input context. Using this simplified input, DoTS improves the inform rate and success rate by 1.09 points and 1.24 points, respectively, compared to the previous state-of-the-art model on MultiWOZ, which is a well-known benchmark.
\end{abstract}

\section{Introduction}
Task-oriented dialogue systems help users achieve their goals in specific domains via conversation. In multi-domain dialogues, the goals are not limited to just one domain; instead, the domain of the conversation changes as the conversation progresses. Thus, the dialogue systems should catch the domain of the current conversation and track the flow of the entire dialogue.

Typical task-oriented dialogue systems use a pipeline architecture that consists of four sequential modules of natural language understanding (NLU), dialogue state tracking (DST), dialogue policy (POL), and natural language generation (NLG). The NLU module extracts important information from the conversations. The DST module tracks the belief state representing constraints on user goals. Furthermore, the output of the DST module is used to query a task-specific database (DB). The POL module determines the next system actions to generate appropriate responses or call APIs. The NLG module generates system responses based on system actions. Recently, researchers have attempted to use methods based on neural dialogue systems, which are trainable in an end-to-end manner. They use the entire dialogue history accumulated through multiple conversational turns for abundant contextual information. However, as the conversation progresses, the dialogue history becomes longer, increasing memory usage and computational costs.

In this paper, we propose the \textbf{Do}main State \textbf{T}racking for a \textbf{S}implified Dialogue System (DoTS), which is an efficient multi-domain task-oriented dialogue system with a simplified input context using the domain state. The novelty of our approach is that DoTS only uses a current user statement instead of the entire dialogue history. To prevent any loss of contextual information, DoTS uses the domain state for the input context, so it can track the flow of conversations from the perspective of domains in a multi-domain dialogue. By removing the entire dialogue history, DoTS uses a memory that is almost constant and less regardless of the length of the conversations; therefore, it can be stably and efficiently trained using long conversations.

DoTS achieves state-of-the-art results without a dialogue history on the standard benchmark datasets MultiWOZ 2.0 and MultiWOZ 2.1, improving the inform rate by 1.09 points and success rate by 1.24 points on MultiWOZ 2.0, and the inform rate by 1.65 points and success rate by 3.68 points on MultiWOZ 2.1. These results show that DoTS performs well without using the entire dialogue history, and that tracking the domain state is helpful for multi-domain dialogue systems.

\begin{figure*}
    \centering
    \includegraphics[width=0.80\textwidth]{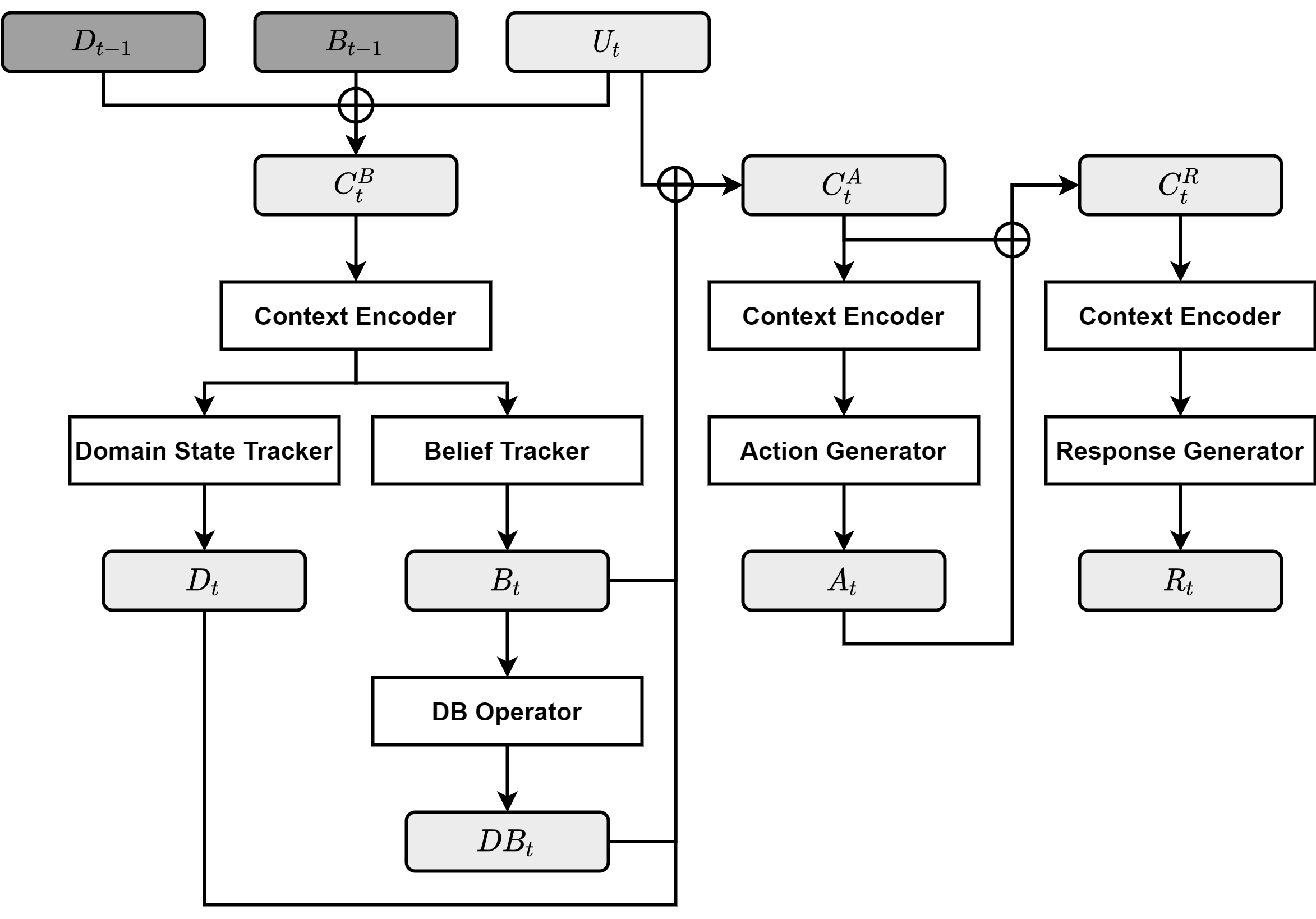}
    \caption{Overview of DoTS. The context encoder is shared between the belief state, system action, and system response. $\oplus$ indicates concatenation.}
    \label{figure1}
\end{figure*}

\section{Related Work}
Existing end-to-end task-oriented dialogue systems consist of NLU, DST, POL, and NLG modules using the entire dialogue history \citep{mehri2019structured, zhang2020task, le2020uniconv, zhang2020probabilistic}.

Recent studies have attempted to use pre-trained language models, such as BERT \citep{devlin2019bert} and GPT-2 \citep{radfordlanguage}, in task-oriented dialogue systems \citep{hosseini2020simple, peng2020soloist}. \citet{peng2020soloist} demonstrated a dialogue system as a large-scale language model instead of a neural pipeline.

\section{Method}
In this section, we describe DoTS, including the architecture, domain state tracking, and context representation. Figure \ref{figure1} presents an overview of DoTS.

\begin{figure*}
    \centering
    \includegraphics[width=0.80\textwidth]{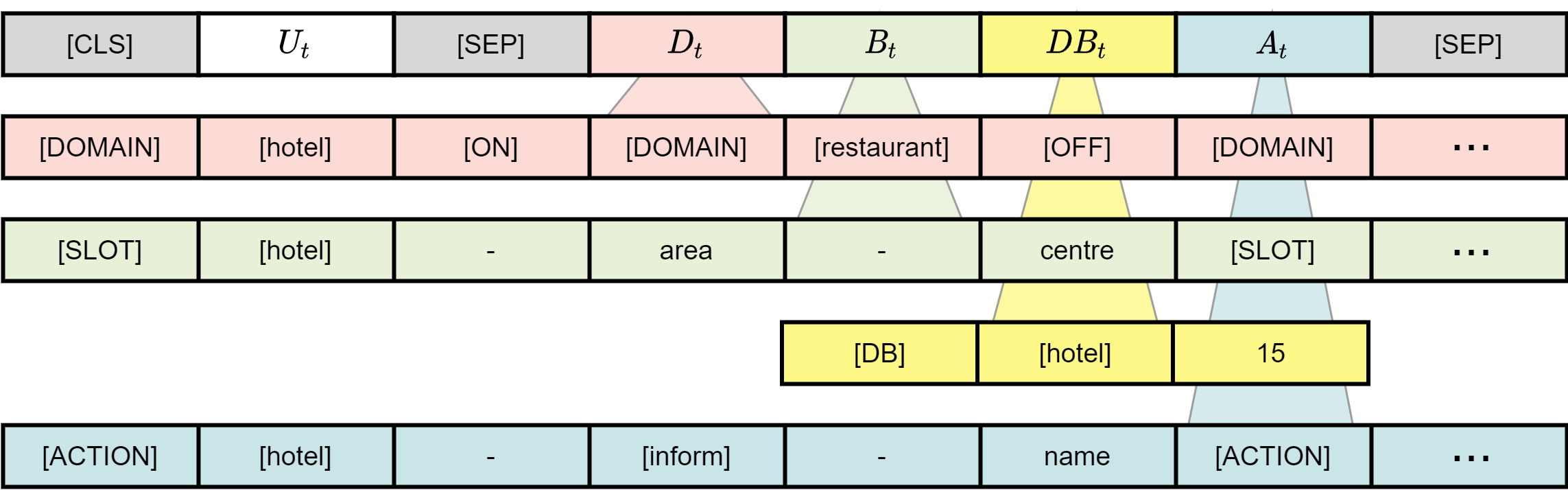}
    \caption{$C^R_t$, the input context for response generation. $D_t$ and $B_t$ involve all domains and all slots, respectively. $DB_t$ includes the number of matched entries of only the activated domains.}
    \label{figure2}
\end{figure*}

\subsection{Neural Pipeline of DoTS}
DoTS consists of a context encoder, domain state tracker, belief tracker, DB operator, action generator, and response generator. The BERT-based context encoder extracts important information from the input context and generate a context output from \texttt{[CLS]} special token. The encoder is shared between the belief state, system action, and system response, and it receives a different input context for each step to calculate the context output. The entire DoTS process is as follows:
\begin{enumerate}
    \item The encoder makes context output $O^B_t$ from input context $C^B_t$ that comprises current user utterance $U_t$, previous domain state $D_{t-1}$, and belief state $B_{t-1}$.
    \begin{align}
        C^B_t &= U_t \oplus D_{t-1} \oplus B_{t-1} \\
        O^B_t &= \texttt{ContextEncoder}(C^B_t)
    \end{align}
    \item Based on $O^B_t$, the domain state tracker classifies new domain state $D_t$, and the belief tracker generates new belief state $B_t$.
    \begin{align}
        D_t &= \texttt{DomainStateTracker}(O^B_t) \\
        B_t &= \texttt{BeliefTracker}(O^B_t)
    \end{align}
    \item The DB operator queries the task-specific DB based on $B_t$ and obtains DB result $DB_t$, indicating the number of matched entries.
    \begin{align}
        DB_t &= \texttt{DBOperator}(B_t)
    \end{align}
    \item The encoder makes context output $O^A_t$ from new input context $C^A_t$ that contains $D_t$ and $B_t$ instead of $D_{t-1}$ and $B_{t-1}$, respectively.
    \begin{align}
        C^A_t &= U_t \oplus D_t \oplus B_t \oplus DB_t \\
        O^A_t &= \texttt{ContextEncoder}(C^A_t)
    \end{align}
    \item The action generator generates system action $A_t$ based on $O^A_t$.
    \begin{align}
        A_t &= \texttt{ActionGenerator}(O^A_t)
    \end{align}
    \item The encoder makes context output $O^R_t$ from new input context $C^R_t$ that additionally contains $A_t$.
    \begin{align}
        C^R_t &= U_t \oplus D_t \oplus B_t \oplus DB_t \oplus A_t \\
        O^R_t &= \texttt{ContextEncoder}(C^R_t)        
    \end{align}
    \item The response generator generates system response $R_t$ based on $O^R_t$.
    \begin{align}
        R_t &= \texttt{ResponseGenerator}(O^R_t)
    \end{align}
\end{enumerate}
We represent the domain state, belief state, DB result, and system action in the form of a sequence of tokens. $\oplus$ indicates concatenation of the sequences, and $t$ indicates the index of conversational turn.

\subsection{Domain State Tracking}
DoTS tracks the domain state in addition to the belief state. The domain state consists of binary values indicating whether domains are activated or not in the current conversation. The domain state tracker classifies the current domain state based on the previous domain state and the current user utterance. The domain state contains the binary values of all domains because it has one or more activated domains. To represent domain state $D_t$ in the form of a sequence of tokens, we substitute the binary values with special tokens \texttt{[ON]} and \texttt{[OFF]}. Furthermore, domains such as \texttt{hotel} and \texttt{taxi} are represented as special tokens. Thus, the domain state is represented as follows:
$$D_t = \texttt{[hotel][ON]...[taxi][OFF]}$$
The domain state enables DoTS to track the flow of the entire dialogue from the perspective of domains. Thus, tracking the domain state can compensate for the loss of contextual information due to neglecting the entire dialogue history.

\begin{table*}
    \centering
    \begin{tabular}{l|c|ccc}
        \Xhline{5\arrayrulewidth} Model & History & Inform$\uparrow$ & Success$\uparrow$ & BLEU$\uparrow$ \\
        \hline SFN+RL \citep{mehri2019structured} & \checkmark & 73.80 & 58.60 & \textbf{16.88} \\
        DAMD+augmentation \citep{zhang2020task} & \checkmark & 76.30 & 60.40 & 16.60 \\
        SimpleTOD \citep{hosseini2020simple} & \checkmark & 84.40 & 70.10 & 15.01 \\
        SOLOIST \citep{peng2020soloist} & \checkmark & 85.50 & 72.90 & 16.54 \\
        \hline DoTS & & \textbf{86.59} & \textbf{74.14} & 15.06 \\
        DoTS (w/o Domain State) & & 84.30 & 68.70 & 15.29 \\
        \Xhline{5\arrayrulewidth}
    \end{tabular}
    \caption{End-to-end evaluation on the MultiWOZ 2.0 test set.}
    \label{table1}
\end{table*}

\begin{table*}
    \centering
    \begin{tabular}{l|c|ccc}
        \Xhline{5\arrayrulewidth} Model & History & Inform$\uparrow$ & Success$\uparrow$ & BLEU$\uparrow$ \\
        \hline UniConv \citep{le2020uniconv} & \checkmark & 72.60 & 62.90 & \textbf{19.80} \\
        LABES-S2S \citep{zhang2020probabilistic} & \checkmark & 78.07 & 67.06 & 18.13 \\
        SimpleTOD \citep{hosseini2020simple} & \checkmark & 85.00 & 70.50 & 15.23 \\
        \hline DoTS & & \textbf{86.65} & \textbf{74.18} & 15.90 \\
        DoTS (w/o Domain State) & & 84.36 & 68.74 & 14.86 \\
        \Xhline{5\arrayrulewidth}
    \end{tabular}
    \caption{End-to-end evaluation on MultiWOZ 2.1 test set.}
    \label{table2}
\end{table*}

\subsection{Context Representation}
Input contexts $C^B_t$, $C^A_t$, and $C^R_t$ are sequences of tokens, including some special tokens. Figure \ref{figure2} presents an example of the input context for response generation, $C^R_t$. For the BERT-based encoder, we use the special tokens \texttt{[CLS]}, which represents the entire sequence, and \texttt{[SEP]}, which separates the sequence into a user utterance and other information. \texttt{-} is the delimiter, which distinguishes the domain, slot, value, and system action.

\section{Experiments}

\subsection{Datasets}
We used the well-known benchmark datasets, MultiWOZ 2.0 \citep{budzianowski2018multiwoz} and MultiWOZ 2.1 \citep{eric2020multiwoz}. These datasets contain approximately 8,000 training, 1,000 validation, and 1,000 test sets, including annotations for end-to-end dialogue systems. MultiWOZ 2.1 is an improved version of MultiWOZ 2.0, where some annotation errors have been corrected.

The evaluation metrics for end-to-end systems are the inform rate, success rate, and BLEU score. The inform rate indicates how many entries provided by the system meet the goals. The success rate additionally measures how many user requests are provided by the system. The BLEU score measures the similarity between the generated response and the true response in the datasets.

\begin{figure}
    \centering
    \includegraphics[width=0.95\columnwidth]{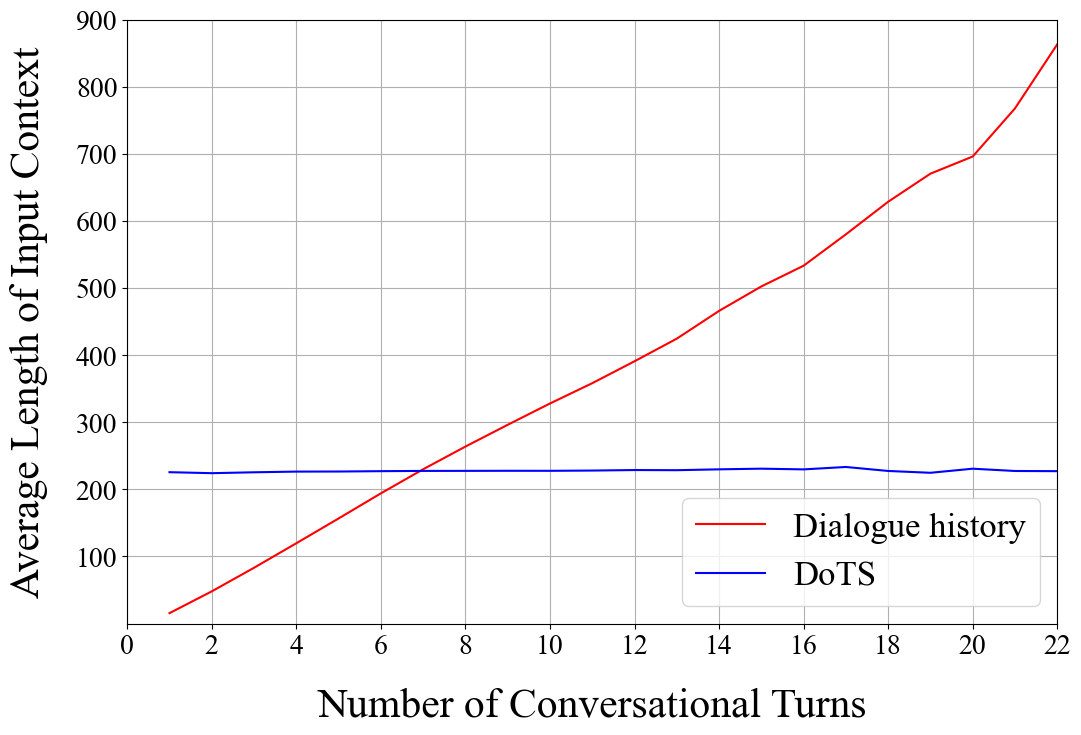}
    \caption{Lengths of the input context of DoTS and entire dialogue history for the number of conversational turns in MultiWOZ.}
    \label{figure3}
\end{figure}

\subsection{Training Details}
We used BERT-base-uncased\footnote{https://github.com/huggingface/transformers} for the context encoder, a fully connected layer for the domain state tracker, and one-layer GRUs \citep{cho2014properties} for the belief tracker, action generator, and response generator. We fine-tuned the pre-trained BERT and trained other modules from scratch. We simultaneously trained all the modules by minimizing the cross entropy losses using an Adam optimizer \citep{DBLP:journals/corr/KingmaB14} with a learning rate of $3\times10^{-5}$, $\beta_1 = 0.9$, and $\beta_2 = 0.999$. We applied an early stopping method; that is, we stopped the training when the sum of inform rate and success rate does not improve over five epochs.

\begin{figure}
    \centering
    \includegraphics[width=0.95\columnwidth]{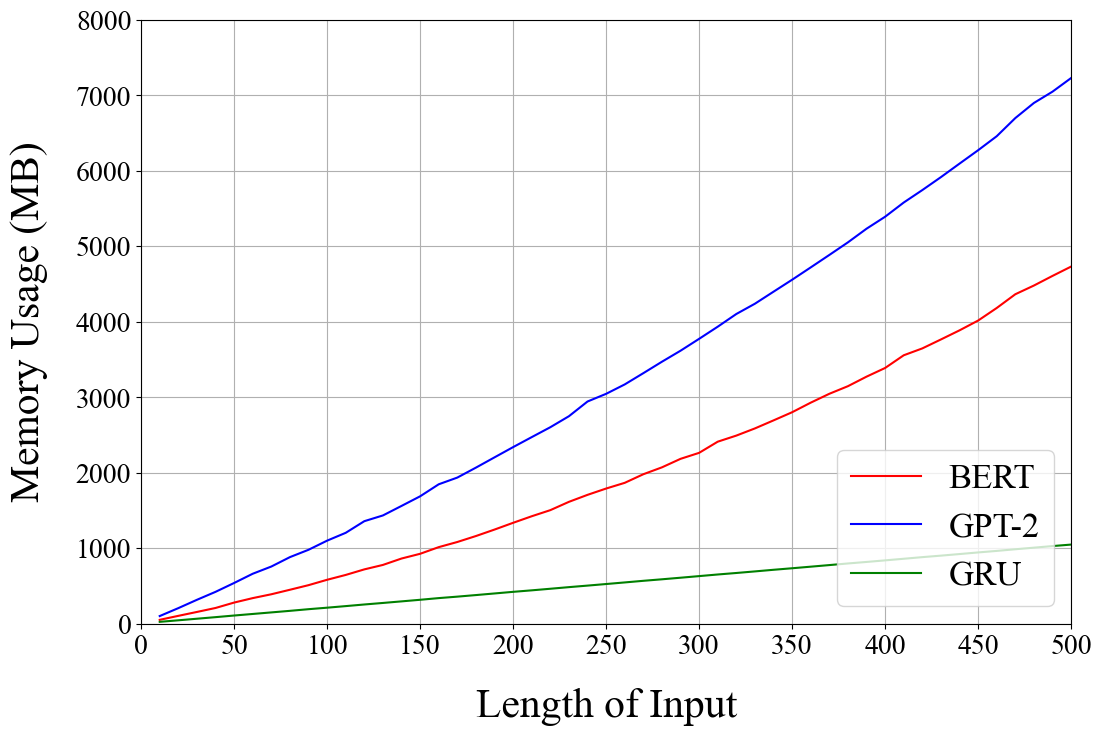}
    \caption{Memory usage of each language model based on input length.}
    \label{figure4}
\end{figure}

\subsection{Results}
The evaluation results on the MultiWOZ 2.0 and MultiWOZ 2.1 test sets, as shown in Table \ref{table1} and Table \ref{table2}, respectively, show that DoTS performs well with only the simplified input context, without using the entire dialogue history. The results also show that tracking the domain state is meaningful for multi-domain dialogue systems. We report the means across 10 runs due to randomness.

We conducted another experiment to confirm that the input context of DoTS is simpler than the entire dialogue history. Figure \ref{figure3} shows a comparison of the lengths for the number of turns. The input context of DoTS has an almost constant length regardless of the number of turns. On the other hand, the dialogue history becomes increasingly longer as the number of turns increases. Figure \ref{figure4} shows a comparison of the memory usage of three language models based on input length. They have 12 layers with hidden size of 768, however, GPT-2 has larger vocabulary. Same increment of input length more increases memory usage on larger model. The memory usage indicates allocated amount of memory on GPU, except the model size, after an input with batch size 8 passes through the language models.

\begin{figure}
    \centering
    \includegraphics[width=0.95\columnwidth]{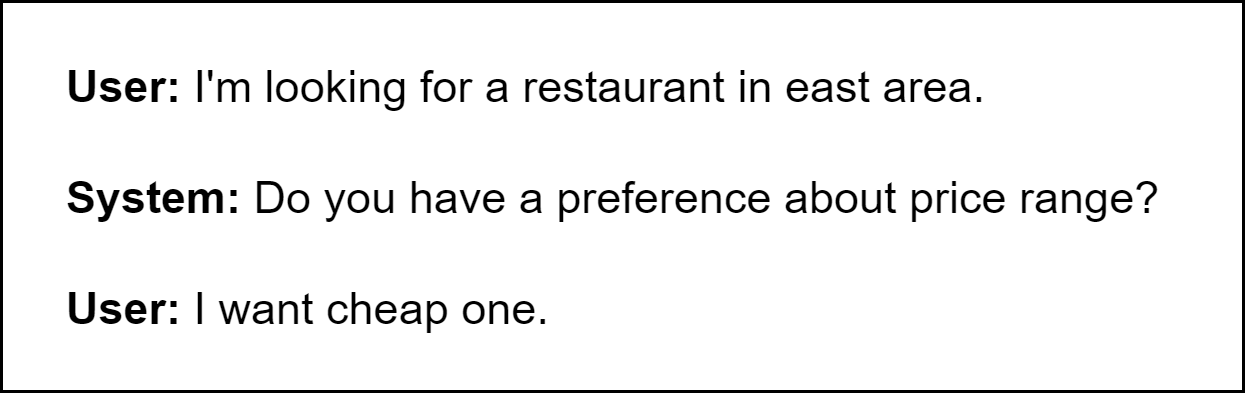}
    \caption{Example of dialogue.}
    \label{figure5}
\end{figure}

\section{Discussion}
Recently, using a pre-trained language model has been a trend in NLP field and has improved performances on various tasks including task-oriented dialogue system.
However, the pre-trained language models such as BERT and GPT-2 have many parameters and accelerate training cost.
Therefore, it is important to summarize input context on large-scale models.

However, summarizing some long text generally causes a loss of contextual information.
Thus, we attempted to use domain state and belief state instead of entire history.
Belief state indicates the constraints of user accumulated through multiple turns, therefore, it has the summarized information from dialogue history.
However, belief state is not perfect summary of history.
Figure \ref{figure5} shows an example of dialogue.
In the above dialogue, the system should look up DB and find a cheap restaurant in east area.
However, the system cannot distinguish that the \texttt{cheap} means cheap restaurant or cheap hotel in multi domain setting without the previous utterances.

By tracking domain state and using it as a part of input context in addition to belief state, the system can follow the flow of dialogue from domain perspective and tackle the above scenario without previous history.

\section{Conclusion}
In this paper, we presented DoTS, a simplified task-oriented dialogue system with domain state tracking. Experimental results demonstrate that task-oriented dialogue systems perform well without the entire dialogue history, and tracking the domain state is helpful for multi-domain dialogue systems.

Using large networks has recently become a trend in natural language processing (NLP) tasks; therefore, large memory and computational costs are required. Thus, the importance of efficiency has increased. We hope that DoTS can inspire the NLP community to explore new, more efficient methods for task-oriented dialogue systems.

DoTS uses some special tokens that are important for specific domains; therefore, it is difficult to transfer it to another domain. As part of our future work, we plan to explore a more universal dialogue system that can be easily transferred to other domains.

\end{document}